\documentclass{article}
\usepackage[papersize={210mm,297mm}, total={160mm,250mm}]{geometry}

\usepackage[square,numbers]{natbib}
\usepackage[utf8]{inputenc} 
\usepackage[T1]{fontenc} 
\usepackage{hyperref} 
\usepackage{url} 
\usepackage{booktabs} 
\usepackage{amsfonts} 
\usepackage{nicefrac} 
\usepackage{microtype} 
\usepackage{lipsum}
\usepackage{fancyhdr} 
\usepackage{graphicx} 
\graphicspath{{media/}} 
\usepackage{amsmath}
\usepackage{nccmath}
\usepackage{tabularx}
\usepackage{multirow}
\usepackage{amssymb}
\usepackage{subcaption}

\usepackage{graphicx}
\usepackage{subcaption}

\usepackage{soul}
\usepackage{color}
\sethlcolor{yellow} 
\usepackage{listings}
\usepackage{siunitx}
\usepackage{longtable}

\usepackage{booktabs}
\usepackage{multirow}

\usepackage[table]{xcolor}

\usepackage{array}
\usepackage{geometry}
\usepackage{ragged2e}
\usepackage[table]{xcolor}

\definecolor{lightgray}{gray}{0.9}

\newcolumntype{P}[1]{>{\RaggedRight\arraybackslash}p{#1}}

\graphicspath{ {images/} }
\title{PRISMA-DFLLM: An Extension of PRISMA for Systematic Literature Reviews using Domain-specific Finetuned Large Language Models

}

\author{
  Teo Susnjak\footnote{\href{mailto:t.susnjak@massey.ac.nz}{Email: t.susnjak@massey.ac.nz}} \\
  School of Mathematical and Computational Sciences \\
  Massey University \\
  Auckland, New Zealand \\
}

\begin{document}
\maketitle

\begin{abstract}
With the proliferation of open-sourced Large Language Models (LLMs) and efficient finetuning techniques, we are on the cusp of the emergence of numerous domain-specific LLMs that have been finetuned for expertise across specialized fields and applications for which the current general-purpose LLMs are unsuitable. In academia, this technology has the potential to revolutionize the way we conduct systematic literature reviews (SLRs), access knowledge and generate new insights. This paper proposes an AI-enabled methodological framework that combines the power of LLMs with the rigorous reporting guidelines of the Preferred Reporting Items for Systematic Reviews and Meta-Analyses (PRISMA). By finetuning LLMs on domain-specific academic papers that have been selected as a result of a rigorous SLR process, the proposed PRISMA-DFLLM (for Domain-specific Finetuned LLMs) reporting guidelines offer the potential to achieve greater efficiency, reusability and scalability, while also opening the potential for conducting incremental living systematic reviews with the aid of LLMs. Additionally, the proposed approach for leveraging LLMs for SLRs enables the dissemination of finetuned models, empowering researchers to accelerate advancements and democratize cutting-edge research. This paper presents the case for the feasibility of finetuned LLMs to support rigorous SLRs and the technical requirements for realizing this. This work then proposes the extended PRISMA-DFLLM checklist of reporting guidelines as well as the advantages, challenges, and potential implications of implementing PRISMA-DFLLM. Finally, a future research roadmap to develop this line of AI-enabled SLRs is presented, paving the way for a new era of evidence synthesis and knowledge discovery.\end{abstract}



\maketitle

\section{Introduction}

\begin{figure}[htbp]
    \centering
        \includegraphics[width=0.6\textwidth]{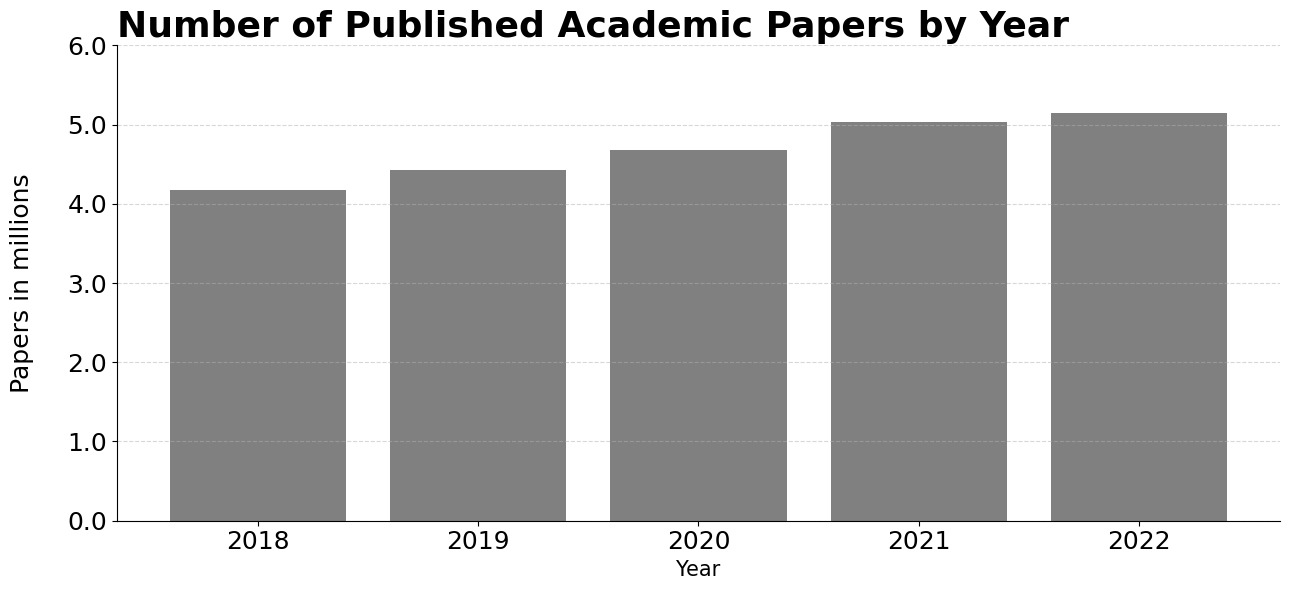}
        \caption{Number of reported \cite{wordsrated2023} published academic papers in the last 5 years.}
        \label{fig:papers}

\end{figure}

The rapid expansion of academic literature across various fields presents a significant challenge for researchers seeking to perform evidence synthesis over the vast body of available knowledge \cite{landhuis2016scientific} (refer to Figure \ref{fig:papers}). Systematic literature reviews (SLRs) have emerged as indispensable tools for evidence-based research, providing comprehensive overviews, synthesizing existing knowledge, and identifying gaps. However, the traditional manual approach to conducting SLRs is not only labor-intensive and resource-draining but also prone to biases. Furthermore, it represents a standalone piece of work that is not easily reusable, incrementally updated, or extended by other researchers as new literature emerges. Given the growing volume of literature, there is an urgent need for more efficient and scalable methods for conducting robust literature reviews and knowledge syntheses.

\begin{figure}[htbp]
    \centering
    \begin{subfigure}[b]{0.48\textwidth}
        \includegraphics[width=\textwidth]{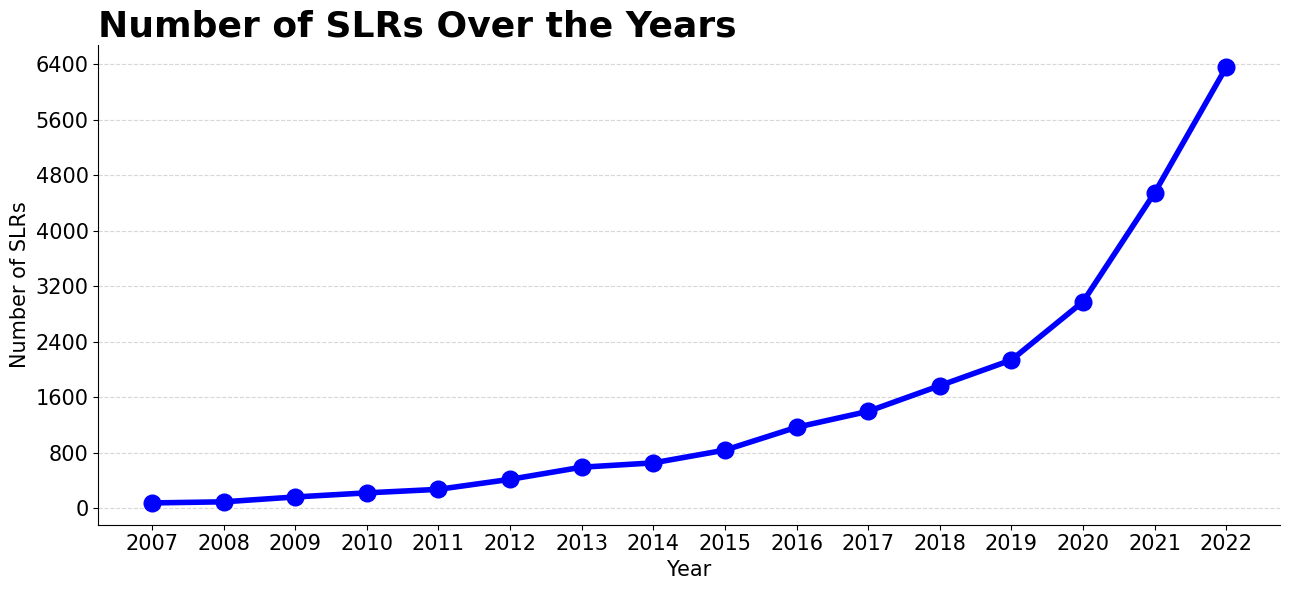}
        \caption{}
        \label{fig:SLRs}
    \end{subfigure}
    \hfill
    \begin{subfigure}[b]{0.48\textwidth}
        \includegraphics[width=\textwidth]{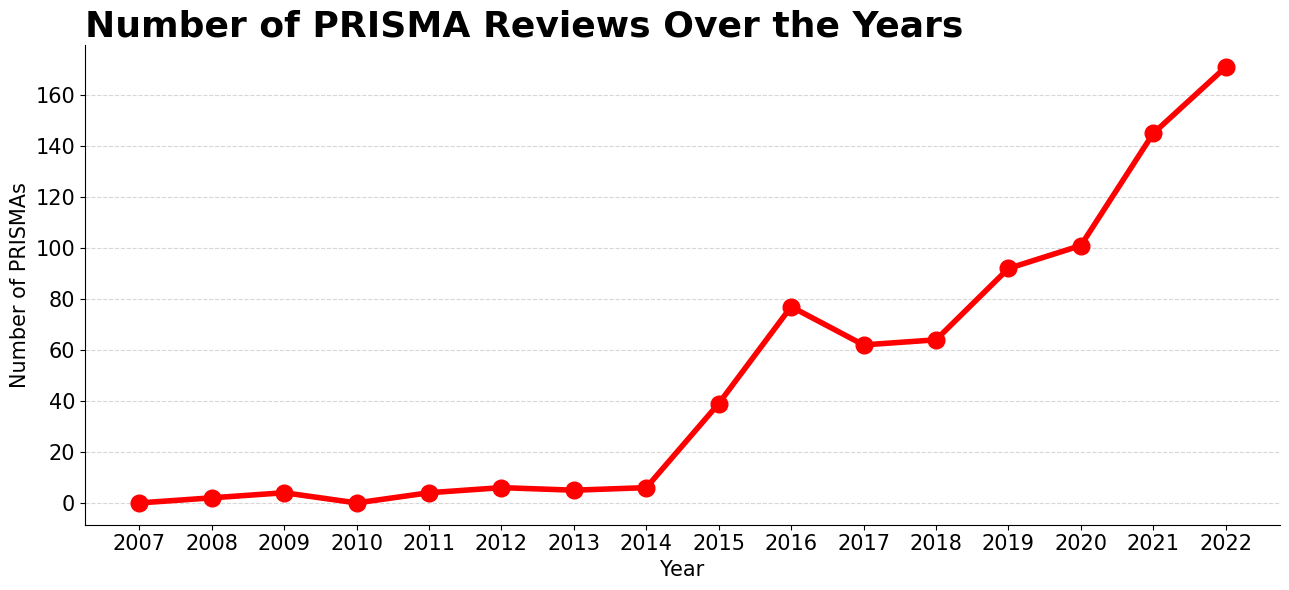}
        \caption{}
        \label{fig:PRISMAs}
    \end{subfigure}
    \caption{Total number of SLRs (a) and PRISMA (b) studies published by year according to Google Scholar.}
    \label{fig:both}
\end{figure}

SLRs are rigorous research methodologies designed to identify, evaluate, and synthesize existing studies on a specific research question. They are highly valued for their structured and comprehensive approach, which aims to minimize bias and promote transparency and replicability. Reporting standards, such as the PRISMA (Preferred Reporting Items for Systematic Reviews and Meta-Analyses) Statement and its recent 2020 extensions \cite{page2021prisma}, provide evidence-based guidelines for achieving high-quality SLRs that maximize transparency and comprehensive reporting. The roadmap \cite{sarkis2021properly} it provides assists authors in accurately describing their research methodology, findings, and planned approach for review protocols. The increasing number of SLRs and PRISMA reviews (Figures \ref{fig:SLRs} and \ref{fig:PRISMAs}) underscores their utility and indicates a need for more efficient strategies to assist researchers.

While the PRISMA framework has greatly contributed to enhancing the transparency and reporting quality of systematic reviews, it is important to acknowledge certain limitations, difficulties, and constraints associated with its implementation.
Conforming to the PRISMA suggestions and conducting evidence syntheses demand a considerable investment of time, effort, and resources\cite{page2021prisma,Kolaski2023}. Researchers need to carry out comprehensive literature searches, apply strict study selection criteria, extract data, and synthesize findings, which can be a resource-intensive process. This poses challenges, particularly for researchers with limited time or funding. Furthermore, once the work is completed, continuous and incremental updates are often not carried out, and a new review in a given field is usually conducted several years later, even by the same team of authors. Subsequent reviews by the same teams only gain minor efficiencies from their previous work, as new research questions require each study to be revisited thoroughly once again.

Despite the rigorous reporting methodology outlined by PRISMA, like any study, systematic reviews are susceptible to bias, errors and selective reporting \cite{page2021prisma,Kolaski2023}. Publication bias, where studies with positive or statistically significant results are more likely to be published, can lead to an overestimation of treatment effects in medical contexts. Additionally, the selective inclusion of studies based on specific criteria can introduce bias and compromise the review's comprehensiveness. \citet{Kolaski2023} note that PRISMA, being a reporting guideline for systematic reviews, is most valuable when consulted during the development of a review rather than as a mere checklist for journal submission. It assesses the completeness of reporting but does not evaluate the quality or performance of the review itself. Therefore, cannot be assumed that strictly following PRISMA guidelines alone guarantees a rigorous systematic review.

Systematic reviews employing PRISMA guidelines aim to objectively synthesise available evidence. However, the interpretation of the evidence and drawing accurate conclusions can be complex \cite{page2021prisma}. 
Additionally, \citet{page2021prisma} note that as both research methodologies and technological advancements continue to evolve rapidly, the PRISMA framework needs to remain up to date with these changes to ensure its relevance and applicability. The authors state that regular updates and adaptations to incorporate emerging methods, such as machine learning or network meta-analyses, are necessary to address the evolving needs of systematic reviews. Given the enormously disruptive effect that the recent advancements and releases of generative AI technologies have had both on academia and beyond, the time has come to consider how PRISMA and systematic literature reviews can be empowered by AI to enable and accelerate future research.

\subsection{Living Systematic Reviews}
Living systematic reviews (LSRs) are a relatively new approach to literature reviews that are designed to provide a more dynamic and up-to-date synthesis of evidence. They are particularly useful for topics where the evidence base is rapidly evolving, and frequent updates are likely to result in changes in effect size or direction. This makes them highly relevant for policy requirements that demand regular updates due to shifting knowledge needs \cite{elliott2017living}.

In an LSR, the literature search and data extraction processes are continually updated to incorporate new evidence as it becomes available. This contrasts with traditional systematic reviews, which provide a snapshot of the evidence at a specific point in time. LSRs, therefore, offer a more current and comprehensive overview of the evidence base, which can be particularly valuable in fast-moving research fields or in response to urgent policy or practice decisions \cite{elliott2017living} requiring new tools and technologies to enable this. The importance of LSRs is increasingly being recognized in the research community where they are regarded to be of most value in fields where there is a high level of uncertainty, and new research is likely to influence the conclusions of the review \cite{elliott2017living}. For instance, the Cochrane Collaboration, a global network of researchers known for producing high-quality systematic reviews, has begun to explore the concept of LSRs in response to the rapidly evolving evidence base in many areas of health care \cite{elliott2017living}. Since conducting an LSR is resource-intensive that requires ongoing search and screening of literature, continuous data extraction and analysis, and regular updates to the review manuscript, entirely new approaches, tools and technologies are needed to enable this.

\subsection{AI and Academic Research Automation}

Recent machine learning advancements in Large Language Model (LLM) developments, and specifically with GPT-class (Generative Pre-trained Transformer) models such as GPT-3.5 and GPT-4 from OpenAI \cite{openai}, have demonstrated unprecedented AI capabilities in natural language understanding and generation of human-like text. These models have exhibited an unprecedented ability to generate human-like text, demonstrating a profound understanding of academic literature across a diverse array of fields. As these models continue to evolve, they are becoming increasingly adept at mitigating the phenomenon of hallucination\footnote{In the context of LLMs, hallucination refers to the generation of outputs or responses that are incorrect, misleading, or unrelated to the given input or prompt.}, thereby enhancing their reliability and potential to assist in academic research where accuracy is paramount. The integration of plugins and web capabilities has further augmented these technologies, enabling them to access and incorporate the latest research findings. 

However, despite these advancements, the most capable and publicly accessible LLM models remain general-purpose and do not yet possess sufficiently specialized and in-depth knowledge across a wide range of disciplines to support the conduct of an SLR. The finetuning of the publicly accessible models (particularly GPT models) for domain-specific knowledge required for SLRs has not been possible, primarily due to the proprietary and confidential nature of the model parameters for these LLMs. Moreover, the computational demands associated with finetuning models comprising billions of parameters for new task capabilities and new knowledge has also until recently has been prohibitively expensive \cite{dettmers2023qlora}.

However, the landscape is rapidly changing with the recent release of several large open-sourced LLMs (examples of the GPT models are Falcon, MosaicML, LLaMA\cite{touvron2023llama}) and the development of techniques that facilitate the finetuning (for example Low-Rank Adaptation (LoRA) \cite{hu2021lora} and Quantized LoRA\cite{dettmers2023qlora}) of LLMs on modest computational and memory resources without compromising accuracy. Work by \citet{dettmers2023qlora} has shown that QLoRA can perform as well as classical full-model finetuning while demonstrating this on the largest publicly available models while performing the finetuning on a single GPU marking a significant shift in the accessibility of LLM finetuning. Thus, these advancements pave the way for the natural evolution towards the proliferation of finetuned domain-specific LLMs across research teams and industries, enhancing these models' expert capabilities and potentially revolutionising how we approach and conduct academic research.

\subsection{Contribution}

Taking advantage of the recent technological progress, this paper proposes both a novel methodology and an AI-integrated PRISMA framework for domain-finetuned LLMs (PRISMA-DFLLM), which combines the power of generative AI language models with the rigorous reporting guidelines of PRISMA. The aim of the proposed reporting extension is to offer effective reporting guidelines for LLM-based SLRs and foster the development of domain-specific LLMs. This will expedite and broaden the reach of research, especially with the growing number of publications and the support of generative AI technologies, ensuring scalability in the research environment.
The main focus of this work is to improve the process of conducting SLRs by addressing the challenges associated with manual methods. The aim is to introduce a more efficient and scalable approach to conducting SLRs, LSRs and other forms of knowledge syntheses.

Additionally, this work proposes a research roadmap for developing LLMs that can be finetuned to support academic research. The roadmap outlines key steps such as selecting domain-specific knowledge, adjusting model parameters, and evaluating model performance. The paper also identifies future research questions that need to be explored, such as how to integrate domain-specific knowledge into LLMs effectively and how to ensure the reliability and validity of finetuned LLMs. 
Finally, this study delves into the benefits and obstacles of developing domain-specific LLMs for research purposes. On one hand, these models can potentially revolutionize how we approach and conduct academic research by automating the literature review process, improving research scalability, and generating unique insights that conventional methods may not uncover. On the other hand, there are a few challenges that need addressing, such as the confidential nature of LLM parameters, the computational demands associated with refining large models, and the importance of thorough evaluation to verify the accuracy and dependability of refined models.

\section{Background}

Over the past few decades, several initiatives have aimed to enhance the quality and transparency of reporting in meta-analyses and systematic reviews. Notable reporting standards and guidelines include the Quality of Reporting of Meta-analyses (QUOROM) \cite{moher1999improving} Statement, the Meta-analyses Of Observational Studies in Epidemiology (MOOSE) \cite{stroup2000meta} guidelines which were superseded by  PRISMA and its recent update PRISMA 2020. The QUOROM Statement, published in 1999, introduced a checklist for meta-analysis reports, covering aspects such as eligibility criteria, search methods, data extraction, and statistical analysis. In 2000, the MOOSE guidelines specifically addressed the reporting of meta-analyses of observational studies, providing recommendations on study design, search strategy, data collection, and study quality assessment.

notably, the revised PRISMA 2020 now acknowledges the importance of methodological quality and risk of bias by incorporating considerations from AMSTAR-2 and ROBIS in its development. 
The newly added reporting of methodological quality and especially the risk of bias in assessment into PRISMA 2020 is a welcome addition with respect to integrations with LLMs and the potential of bias.

In addition to PRISMA, two other significant approaches \cite{Kolaski2023}  that support evidence-based practice with a rigorous methodology and transparency are AMSTAR-2 \cite{shea2017amstar}  and ROBIS \cite{whiting2016robis}. AMSTAR-2 provides a comprehensive checklist for assessing methodological quality, while ROBIS focuses on evaluating the risk of bias within systematic reviews. In contrast, PRISMA 2020 is a reporting guideline that emphasizes complete and transparent reporting of systematic reviews. Both AMSTAR-2 and ROBIS enhance the reviewers' ability to appraise the overall quality and validity of systematic reviews. Notably, the revised PRISMA 2020 now incorporates considerations from AMSTAR-2 and ROBIS to acknowledge the importance of methodological quality and risk of bias. This inclusion of methodological quality and risk of bias reporting in PRISMA 2020 is a welcome addition that is particularly relevant to integrations with LLMs and potential bias.

\subsection{The PRISMA 2020 Statement}

PRISMA is a guideline based on evidence that outlines the essential items required for reporting systematic reviews and meta-analyses \cite{moher2009preferred}. Created in 2009, PRISMA aims to improve the quality and transparency of systematic reviews by ensuring researchers report crucial information necessary for the critical evaluation and replication of their work. Although it was initially intended for reviews that assess randomized trials, PRISMA has been widely adopted and can serve as a basis for reporting systematic reviews of other study designs, such as observational studies, diagnostic accuracy studies, and qualitative research.

The PRISMA statement was updated in 2020 to reflect the latest developments and challenges in conducting systematic reviews \cite{page2021prisma}. The 2020 version includes a checklist and a flow diagram, like its predecessor. However, the updated checklist now includes 27 items that cover essential components of a systematic review, such as the title, abstract, introduction, methods (such as eligibility criteria, search strategy, study selection, data extraction, and data synthesis), results, discussion, and funding. Each checklist item provides guidance to researchers on what to include in their systematic review report. The flow diagram, also known as the PRISMA flow diagram, visually represents the study selection process, making it easier for readers to understand the flow of studies from identification to the final included studies.

PRISMA has had a significant impact on the field of systematic reviews and meta-analyses since its introduction. It has become widely recognized and endorsed by leading journals and organizations in the field of evidence-based medicine. Researchers, reviewers, and journal editors often refer to PRISMA as a critical tool for improving the transparency, rigor, and completeness of systematic reviews. Adhering to PRISMA guidelines facilitates the assessment of the risk of bias and the replication of studies, enabling the integration of evidence into practice and policy-making.

The PRISMA 2020 update demonstrates a continued commitment to improving the reporting standards in systematic reviews, further enhancing its applicability and usability \cite{page2021prisma}. As research methodologies and technological advancements continue to evolve rapidly, the PRISMA framework must remain up to date with these changes to ensure its relevance and applicability. This commitment to continuous improvement underscores the importance of PRISMA in advancing evidence-based practice and decision-making.

\subsection{Extensions to PRISMA}

There is already a well-established precedent for customizing the original PRISMA guidelines. Diverse research and methodological challenges have led to the proposal of several extensions to the PRISMA framework. These extensions augment the versatility and adaptability of the PRISMA guidelines, empowering researchers to conduct and report systematic reviews in specific domains. By tailoring guidance to address context-specific needs, these extensions bolster transparency, reproducibility, and the overall efficacy of systematic review methods. The extensions discussed in this section include PRISMA-P (Protocols), PRISMA-ScR (Scoping Reviews), PRISMA-NMA (Network Meta-Analyses), PRISMA-IPD (Individual Patient Data), PRISMA-Harms (Harms Reporting), and PRISMA-RR (Rapid Reviews). Each extension equips researchers with a comprehensive checklist and elucidation of critical components, thereby optimizing the quality, transparency, and comprehensibility of systematic reviews within distinct research contexts.

\paragraph*{PRISMA-P (for Protocols)}
PRISMA-P \cite{moher2015preferred} is an extension of the PRISMA framework that specifically focuses on the development and reporting of systematic review protocols. A systematic review protocol outlines the objectives, methods, and analysis plan that will be followed in a systematic review. It serves as a blueprint for conducting the review and ensures transparency, reproducibility, and consistency in the review process. The PRISMA-P extension provides a checklist and explanation for the key items that should be included in a systematic review protocol, covering aspects such as the rationale, eligibility criteria, search strategy, data extraction, and synthesis methods. By adhering to the PRISMA-P guidelines, researchers can enhance the quality and transparency of their systematic review protocols, facilitating a better understanding and assessment of the planned review.

\paragraph*{PRISMA-ScR (for Scoping Reviews)}
Scoping reviews aim to map the literature on a particular topic or research area and provide an overview of the available evidence. Unlike systematic reviews, scoping reviews are typically broader in scope and focus on identifying the main concepts, theories, sources, and gaps in the existing literature. The PRISMA-ScR \cite{tricco2018prisma} extension adapts the original PRISMA guidelines to the unique characteristics of scoping reviews, providing a checklist and explanation to guide researchers in conducting and reporting scoping reviews. It covers key aspects such as the research question, search strategy, study selection process, data extraction, and presentation of findings. By adhering to the PRISMA-ScR guidelines, researchers can enhance the transparency and rigor of their scoping reviews, enabling better understanding and utilization of the synthesized evidence.

\paragraph*{PRISMA-NMA (for Network Meta-Analyses)}

Network meta-analysis (NMA), also known as multiple-treatments meta-analysis or indirect treatment comparison, is a statistical method that allows for the simultaneous comparison of multiple interventions in a single analysis. It enables the estimation of relative treatment effects even in the absence of head-to-head comparisons between all treatments. The PRISMA-NMA \cite{hutton2015prisma} extension provides guidelines for the reporting of network meta-analyses, ensuring transparency and clarity in the reporting of methods, results, and interpretations. The checklist covers key aspects such as the study design, search strategy, study selection criteria, data extraction, risk of bias assessment, statistical analysis methods, and presentation of results. By following the PRISMA-NMA guidelines, researchers can improve the quality and comprehensibility of their network meta-analyses, facilitating the synthesis and interpretation of evidence on comparative treatment effects.

\paragraph*{PRISMA-IPD (for Individual Patient Data)}
Meta-analyses that utilize individual patient data (IPD) provide a more detailed and potentially accurate analysis compared to those relying on aggregate data. IPD allows for the examination of patient-level characteristics and the exploration of treatment effects across subgroups. The PRISMA-IPD \cite{stewart2015preferred} extension focuses on reporting guidelines for systematic reviews and meta-analyses that use individual patient data. The checklist covers items such as the study design, data collection methods, participant characteristics, outcomes of interest, statistical methods used for the analysis, and interpretation of results. By adhering to the PRISMA-IPD guidelines, researchers can enhance the transparency, accuracy, and comparability of their systematic reviews and meta-analyses based on individual patient data.

\paragraph*{PRISMA-Harms (for Harms Reporting)}
The PRISMA-Harms \cite{zorzela2016prisma} extension focuses on the comprehensive reporting of harmful outcomes in systematic reviews and meta-analyses. While systematic reviews traditionally focus on the effectiveness and benefits of interventions, it is equally important to examine and report the potential harms associated with those interventions. PRISMA-Harms provides guidelines to systematically identify, extract, and report data on adverse events and harms in the included studies. The extension emphasizes the need for transparency, completeness, and consistency in reporting adverse events and facilitates a more balanced assessment of the benefits and risks associated with interventions.

\paragraph*{PRISMA-RR (for Rapid Reviews)}

Rapid reviews are a form of knowledge synthesis that aims to produce timely information while streamlining or omitting certain components of the systematic review process. The PRISMA-RR \cite{stevens2018developing} extension provides a reporting checklist specifically tailored to the unique characteristics and requirements of rapid reviews. It covers key aspects such as the review question, search strategy, study selection process, data extraction, and synthesis methods. By following the PRISMA-RR guidelines, researchers can ensure transparency and rigor in reporting rapid reviews, enabling readers to better understand the strengths and limitations of these expedited knowledge syntheses.

\subsection{AI-Enabled SLRs: Current State and Future Directions}

The surge in academic literature has led to the integration of AI technologies in systematic reviews, aiming to streamline the review process and enhance efficiency \cite{o2019question, tsafnat2014systematic}. These technologies have been utilized in various stages of the review process, including literature screening and data extraction from included studies. However, the implementation of AI in systematic reviews is not without challenges, such as the need for extensive labeled data for AI model training and the complexity of automating subjective tasks like quality assessment \cite{marshall2019toward}.

Recent studies have explored the potential of machine learning algorithms and they have shown promise in reducing the workload of reviewers during citation screening \cite{o2019question}. Similarly, text mining techniques have been employed for data extraction from included studies, although the complexity and variability of academic literature data present significant challenges \cite{tsafnat2014systematic}. Knafou et al. \cite{knafou2023ensemble} demonstrated the efficacy of deep learning language models in classifying COVID-19-related publications, suggesting the potential of AI in supporting epidemiological curation and review. Forsgren et al. \cite{forsgren2023use} discussed the utility of text-mining functions in facilitating the screening process for topics with diffuse conceptual boundaries, highlighting the potential of AI in improving the workflow of comprehensive reviews. Muller et al. \cite{muller2023effect} proposed a retrospective study to quantify the effect of machine learning adoption on resource use and time-to-completion in systematic reviews. The study underscores the potential benefits of machine learning in evidence synthesis and highlights the need for further research to address concerns about quality and automation.

However, limitations still exist, and further research is needed to address concerns about quality, automation, and the current limitations of various AI models in evidence synthesis. Qureshi et al. \cite{qureshi2023chatgpt} explored the potential of ChatGPT \cite{openai}, a general-purpose LLM developed by OpenAI, in aiding systematic reviews. While the model showed promise in certain areas, the study highlighted the need for further exploration and testing to understand its current limitations and capacity in the context of evidence synthesis. Only very recently, are we beginning to witness the emergence of domain-specific LLMs, which were created as a result of pretraining on large datasets requiring enormous compute resources. In the domain of healthcare, Singhal et al. \cite{singhal2022large, singhal2023towards} demonstrated the potential of LLMs in medical question answering. Their research on Med-PaLM 2 showcased remarkable advancements towards physician-level performance in this domain, along with the introduction of benchmarks and evaluation frameworks specific to LLMs in medical question answering. Additionally, Stanford CRFM developed PubMedGPT 2.7B \cite{bolton2022stanford}, a language model exclusively trained on biomedical abstracts and papers, which achieved impressive results across various biomedical NLP tasks \cite{bolton2022stanford}. Similarly, in the field of finance, Wu et al. \cite{wu2023bloomberggpt} presented BloombergGPT, a 50 billion parameter language model trained on a diverse range of financial and general-purpose data. This model surpassed existing models in financial tasks while maintaining strong performance on general-purpose benchmarks, highlighting the successful training of LLMs on both domain-specific and general data sources. Meanwhile, Lehman et al. \cite{lehman2023we} investigated the suitability of LLMs trained on general web text for specialized and safety-critical domains like clinical text through extensive analysis of 12 language models, and their study reveals that relatively small specialized clinical models outperform larger LLMs trained on general text, even when finetuned with limited annotated data. Their findings suggest the viability of finetuning smaller LLMs for domain-specific tasks which may be suitable for SLRs.

Despite these advancements, there is a pressing need for an integrated AI-based guideline that can fully support the conduct of systematic reviews, and there is specifically a gap in leveraging LLMs to automate parts of the review process and generate insights directly from the literature \cite{Brown2020}. This approach could not only save time and resources but also enhance the comprehensiveness and quality of systematic reviews. The recent advancements in AI, particularly in the development of LLMs, underscore the potential of developing such an integrated AI framework and beginning to experiment with customized LLMs that have been finetuned for domain-specific knowledge which holds promise for revolutionizing the conduct of systematic reviews.

\subsection{AI Advances in LLM Finetuning}

The advent of LLMs has revolutionized the field of natural language processing, offering unprecedented capabilities in understanding and generating human-like text. However, the sheer size of these models presents significant challenges in terms of computational cost and memory footprint, especially during the finetuning process. To address these challenges, a range of Parameter-Efficient Fine-Tuning (PEFT) strategies have been developed \cite{chen2023parameter}. These strategies aim to update a small subset of parameters while keeping the rest of the model frozen, thereby achieving comparable performance to full finetuning while significantly reducing computational cost and memory footprint.

One of the pioneering PEFT strategies is Low-Rank Adapters (LoRA). LoRA introduces low-rank matrices into each layer of the pre-trained model. These matrices are the only parameters that are updated during finetuning, while the original pre-trained parameters are kept frozen. This approach significantly reduces the number of parameters that need to be updated, making finetuning more efficient and less prone to overfitting. However, it's worth noting that LoRA may not offer the same level of flexibility as full finetuning, as it can only modify the model's behavior in a limited way \cite{hu2023llm}.

Building on the concept of LoRA, Quantized LoRA (QLoRA) was developed to further reduce memory usage by quantizing the pre-trained model's weights. This involves representing the weights with a smaller number of bits, which can significantly reduce the memory footprint of the model. QLoRA also introduces several other innovations to save memory without sacrificing performance, such as double quantization and paged optimizers. However, this approach can be more complex to implement and may introduce additional computational overhead \cite{hu2023llm}. Adapters, for instance, introduce small, task-specific parameter matrices into the model, which are trained while the original parameters are frozen \cite{wang2022adamix}. This approach allows for task-specific adaptations without the need to finetune the entire model, thereby maintaining the benefits of the original pre-training while adapting the model to new tasks. Compared to LoRA and QLoRA, Adapters can offer greater flexibility as they allow for task-specific modifications. However, the introduction of task-specific parameters can increase the complexity of the model and may require more careful management of the training process to avoid overfitting. Prefix tuning, on the other hand, adds a task-specific prefix to the input sequence, which can modify the model's behavior in a more flexible way than LoRA or Adapters \cite{gui2023g}. This approach can be particularly effective for tasks that require a significant change in the model's behavior, as the prefix can guide the model towards the desired output. Compared to LoRA and QLoRA, prefix tuning can offer greater flexibility and can be more effective for tasks that require a significant change in the model's behavior. However, it may also require more careful tuning of the prefix and can potentially introduce more computational overhead due to the need to process the additional input.

PEFT strategies offer a promising approach to finetuning LLMs in a more computationally and memory-efficient manner. However, the effectiveness of these strategies can depend on the specific method used and the characteristics of the task. As such, it is crucial to carefully consider the trade-offs between different PEFT strategies when finetuning LLMs for specific applications. Future research in this area is likely to yield even more efficient and effective strategies for finetuning LLMs.

In order to underscore the feasibility of finetuning, \citet{zhou2023lima} recently concluded that only limited instruction finetuning data is necessary to teach models to produce high-quality output which they demonstrated by training LIMA, a 65B parameter LLaMa language model, with only 1,000 carefully curated prompts and responses, without any reinforcement learning or human preference modeling. While the authors note that almost all knowledge in an LLM is learned during pretraining where general-purpose representations are learned from raw text, it remains possible to encode additional knowledge and necessary downstream behaviours necessary for conducting an SLR with only a small finetning dataset comprising target academic papers.

\section{Outline of Additional PRISMA Reporting Components}

The proposed framework aims to augment the existing PRISMA methodology for SLRs by explicitly incorporating several key components related to LLM finetuning. These additional components address the reporting topics of the finetuning dataset, technical LLM finetuning details, and the evaluation of the finetuned LLM. By integrating these components, the PRISMA-DFLLM methodology expands upon the established PRISMA guidelines without replacing any existing ones. The first step of the proposed guideline is to conduct a traditional PRISMA search criteria for paper collection and filtering. This ensures that the systematic and comprehensive nature of the original PRISMA framework is maintained, resulting in a rigorous selection of relevant literature that is essential and reportable. Subsequently, the identified papers form the basis for the subsequent steps in the PRISMA-DFLLM methodology.

\subsection{Reporting the Finetuning Dataset Details}

The dataset used for finetuning LLMs is of paramount importance and necessitates meticulous processing and preparation and is constructed from the pertinent academic papers. During the data preparation phase, the raw text extracted from these papers undergoes cleaning and preprocessing to ensure uniformity and facilitate smooth processing by the model. This involves tasks such as removing or replacing specific characters, addressing encoding issues, and standardizing formats like dates and numbers.Furthermore, the metadata associated with each paper, encompassing details such as authors, publication date, journal, and keywords, is seamlessly integrated with the paper's text. This integration enables the LLM to discern connections between content and metadata, thereby enhancing its comprehension of the paper and its context. It is crucial to explicitly outline the strategy employed in constructing the dataset, including any automated or manual steps taken to represent information from the academic papers.

In addition to the dataset construction strategy, it is imperative to clearly specify the format of the input and output data for the LLM, along with any domain-specific preprocessing steps and text encoding methods utilized. If supplementary datasets were incorporated for instruct-finetuning or to augment the LLM's general domain knowledge, these should be reported as well. Another important aspect is reporting the size attributes of the final finetuning dataset, as this information impacts the LLM's ability to generalize and the computational resources required for training. 

Since the dataset used for finetuning LLMs necessitates meticulous processing and preparation a structured approach needs to be followed and communicated. Cleaning and preprocessing steps ensure a uniformly structured dataset, while the incorporation of metadata enhances the model's comprehension of the paper and its context. Therefore it is crucial to provide detailed reporting of the dataset construction strategy, format specifications, and any additional datasets utilized to ensure reproducibility and a comprehensive understanding of the LLM's training process and any extensions of PRISMA need to set guidelines that accommodate this.

\subsection{Reporting the LLM Finetuning Process Details}

The selection of a suitable base LLM model for finetuning is of utmost importance in the LLM finetuning process. Different LLM models exhibit variations in terms of their architecture, capacity, and performance on language tasks. When choosing a base LLM model, it is essential to consider these factors and evaluate their alignment with the goals and requirements of the finetuning task. In the context of the PRISMA-DFLLM framework, considerations of various base LLM models, their suitability and the assessment their performance need to be communicated. This ought to be reported since options range from raw LLM models, which have been trained on extensive text corpora but are yet to be finetuned for downstream tasks, to models that have already undergone a degree of finetuning for specific instructions or domains.

The choice of a base LLM model depends on factors such as its architectural features, capacity to capture complex relationships in data, and its performance on language tasks relevant to the finetuning objective. Before commencing the finetuning process, it is essential to understand the status of the chosen base LLM model—whether it is raw or has already undergone some degree of finetuning, and to report this. Raw models offer a broader understanding of language patterns but lack domain-specific knowledge. On the other hand, models that have undergone prior finetuning for specific instructions or domains might already have some domain-specific knowledge embedded which assists the subsequent finetuning. Evaluating the suitability of a base model in terms of its rawness or prior finetuning ensures that the finetuning process aligns with the specific requirements of the task at hand.

Once an appropriate base LLM model has been selected, the finetuning process involves training the model on a specific academic corpus. This is achieved through techniques like adjusting hyperparameters such as learning rate, batch size, and the number of training epochs. An optimization algorithm, such as Adam or Stochastic Gradient Descent, is then utilized to update the model's parameters based on a loss function. The goal is to minimize errors and improve the accuracy of predictions. To prevent overfitting, techniques like dropout, weight decay, or early stopping can be used during the finetuning process. Post-processing steps, such as temperature scaling, can further optimize the LLM's performance by controlling the level of randomness in the generated text outputs. Once the finetuning process is complete, the final LLM model is prepared for deployment by packaging it along with any necessary support files, such as tokenizers or preprocessing tools. This is done in a format that allows researchers to easily load and utilize the model for their intended tasks.

The choice of a suitable base LLM model is therefore consequential in the finetuning process, considering factors such as architectural features, model capacity, and performance on language tasks that need to be reported. Assessing whether the model is in a raw form or has undergone prior finetuning assists in aligning the finetuning process with the specific requirements of the task. The subsequent finetuning process involves adjusting hyperparameters, employing optimization algorithms, and applying techniques to prevent overfitting, all of which ought to be documented as part of the extended PRISMA guideline. Post-processing steps can optimize the model's performance, and the final model is prepared for deployment by packaging it with the necessary support files. Clear reporting of these steps ensures transparency as well as a potential for reproducibility.

\subsection{Reporting the Evaluation of Finetuned LLMs Details}

The evaluation of the finetuned PRISMA-DFLLM model is a multifaceted process designed to provide a robust and comprehensive assessment of its performance. This process is tailored to the model's intended use case, ensuring that the evaluation metrics match the specific tasks the model is designed to perform and that overall it achieves alignment\footnote{Alignment, in the context of LLMs, pertains to the degree of concordance or harmony between the generated outputs of the model and the intended or expected outputs.}. In the context of information retrieval tasks, which are central to SLRs, metrics such as accuracy, precision, recall, and F1-score can be employed. For instance, precision (the proportion of retrieved documents/information that are relevant) and recall (the proportion of relevant documents/information that are retrieved) provide a balanced view of the model's performance in identifying relevant literature. The F1-score, the harmonic mean of precision and recall, gives an overall performance metric.

When the model is used for document summarization tasks, the ROUGE metric can be utilized. This metric compares the overlap of n-grams, word sequences of n words, between the generated summaries and human-written abstracts. For example, a high ROUGE-2 score would indicate a significant overlap of two-word sequences between the model's output and the reference summary, suggesting a high-quality summary.

Human evaluation can also be a valuable part of the evaluation process. It provides a qualitative assessment of factors such as coherence, completeness, and fidelity to the original paper. In generative tasks, human evaluators assess the model's responses for coherence, relevance to the prompt, novelty, and factual accuracy. For instance, evaluators might rate the model's responses on a Likert scale for these factors, providing a more nuanced understanding of the model's performance. Comparative evaluations form another key component of a possible suite of assessments. Here, the performance of the finetuned model is compared against baseline models, such as the original LLM before finetuning or other LLMs finetuned on different corpora. This comparison helps to quantify the added value of the finetuning process and the specific corpus used.

Ensuring the stability and reproducibility of the evaluations is paramount. This can be achieved by running the model multiple times with different random seeds and varying the dataset and the model's initial parameters. Techniques such as data splitting into training, validation, test sets, and cross-validation can be used to enhance the reliability of the evaluations. Additionally, a qualitative analysis of the model's outputs can be conducted, examining case studies of both successful and less successful examples. Errors made by the model  can be categorized and analyzed, providing valuable insights for potential improvements. For example, if the model consistently struggles with a certain type of prompt, adjustments to the finetuning process to better handle such prompts can be made.

There are a raft of possible evaluation metrics that can be applied to finetuned LLMs. Each has their strengths. Given the central importance of a finetned LLM in the proposed SLR process, a robust evaluation and comprehensive process covering quantitative metrics, comparative assessments, reproducibility checks, and qualitative analyses need to be performed and reported. A thoroughly documented approach ensures transparency and reliability of the LLM-based SLRs that deliver high-quality outputs meeting the requirements of SLR processes.

\subsection{Reporting the Considerations of Ethical and Legal Aspects}

The process of finetuning LLMs on a corpus of academic papers necessitates a careful navigation of ethical, legal, and privacy considerations. The potential harm that could arise from the model's outputs, such as the propagation of biases, generation of inappropriate content, or violation of privacy, is a concern that needs to be considered \cite{sheng2021societal}. 

A primary legal concern is compliance with copyright laws. These laws restrict the reproduction, distribution, and public display of copyrighted works, including substantial portions of a work, even if they are transformed. For instance, finetuning a language model on copyrighted text could be viewed as a transformative use, potentially falling under the legal doctrine of "fair use" in certain jurisdictions like the United States. However, the application of fair use is subjective and evaluated based on several factors, including the purpose and nature of the use, the characteristics of the copyrighted work, the amount and substantiality of the portion used, and the impact on the potential market for the copyrighted work. To mitigate the risk of copyright infringement, it is advisable to seek permission from copyright holders, use open-access academic papers when possible, or consult with legal experts \cite{sheng2021societal}. These considerations ought to be reported as part of the SLR.

Data privacy is another significant consideration, especially when handling sensitive information.
For instance, in fields like medical research, academic papers may contain sensitive patient data. In such cases, it is crucial to ensure compliance with data protection laws and ethical guidelines. This process should be documented, and measures should be taken to protect this information during the finetuning process and especially if the intention is to make the finetuned available to the public. 

Lastly, the finetuning process can introduce or amplify biases present in the training data. As \citet{sheng2021societal} discuss, language generation applications like LLMs can exhibit a variety of biases and consequently it is important to monitor and mitigate these where possible. For example, techniques such as bias mitigation algorithms or fairness-aware machine learning methods could be employed. The ethical application of the PRISMA-DFLLM framework requires a commitment to responsible AI use, respect for intellectual property rights, and a proactive approach to monitoring and mitigating potential biases and reporting guidelines of these issues need to accommodate these aspects.

\section{Extended PRISMA Reporting Guidelines}

In light of the previous section, the extension to the original PRISMA 2020 checklist includes several new categories. This section presents a proposed extension to the PRISMA 2020 checklist, tailored specifically for studies involving the finetuning and application of LLMs for conducting SLRs. The following subsection presents a reporting checklist that covers the specifics of the finetuning dataset preparation, the finetuning process of the LLM, the evaluation of the model's performance, and the ethical and legal considerations associated with the use of LLMs.

\subsection{The PRISMA-DFLLM Checklist}

The revised checklist below highlights new items.

\begin{enumerate}
    \item \textbf{TITLE}
    \begin{itemize}
        \item[1.] Title: Identify the report as a systematic review.
    \end{itemize}
    
    \item \textbf{ABSTRACT}
    \begin{itemize}
        \item[2.] Abstract: See the PRISMA 2020 for Abstracts checklist.
    \end{itemize}
    
    \item \textbf{INTRODUCTION}
    \begin{itemize}
        \item[3.] Rationale: Describe the rationale for the review in the context of existing knowledge.
        \item[4.] Objectives: Provide an explicit statement of the objective(s) or question(s) the review addresses.
    \end{itemize}
    
    \item \textbf{METHODS}
    \begin{itemize}
        \item[5.] Eligibility criteria: Specify the inclusion and exclusion criteria for the review and how studies were grouped for the syntheses.
        \item[6.] Information sources: Specify all databases, registers, websites, organisations, reference lists and other sources searched or consulted to identify studies. Specify the date when each source was last searched or consulted.
        \item[7.] Search strategy: Present the full search strategies for all databases, registers and websites, including any filters and limits used.
        \item[8.] Selection process: Specify the methods used to decide whether a study met the inclusion criteria of the review, including how many reviewers screened each record and each report retrieved, whether they worked independently, and if applicable, details of automation tools used in the process.
        \item[9.] Data collection process: Specify the methods used to collect data from reports, including how many reviewers collected data from each report, whether they worked independently, any processes for obtaining or confirming data from study investigators, and if applicable, details of automation tools used in the process.
        \item[10] \textbf{Data items:}
        \begin{itemize}
            \item[10a.] List and define all outcomes for which data were sought. Specify whether all results that were compatible with each outcome domain in each study were sought (e.g. for all measures, time points, analyses), and if not, the methods used to decide which results to collect.
            \item[10b.] List and define all other variables for which data were sought (e.g. participant and intervention characteristics, funding sources). Describe any assumptions made about any missing or unclear information.
        \end{itemize}
        \item[11.] Study risk of bias assessment: Specify the methods used to assess risk of bias in the included studies, including details of the tool(s) used, how many reviewers assessed each study and whether they worked independently, and if applicable, details of automation tools used in the process.
        \item[12.] Effect measures: Specify for each outcome the effect measure(s) (e.g. risk ratio, mean difference) used in the synthesis or presentation of results.
        \item[13] \textbf{Synthesis methods:}
        \begin{itemize}
            \item[13a.] Describe the processes used to decide which studies were eligible for each synthesis (e.g. tabulating the study intervention characteristics and comparing against the planned groups for each synthesis (item \#5)).
            \item[13b.] Describe any methods required to prepare the data for presentation or synthesis, such as handling of missing summary statistics, or data conversions.
            \item[13c.] Describe any methods used to tabulate or visually display results of individual studies and syntheses.
            \item[13d.] Describe any methods used to synthesize results and provide a rationale for the choice(s). If meta-analysis was performed, describe the model(s), method(s) to identify the presence and extent of statistical heterogeneity, and software package(s) used.
            \item[13e.] Describe any methods used to explore possible causes of heterogeneity among study results (e.g. subgroup analysis, meta-regression).
            \item[13f.] Describe any sensitivity analyses conducted to assess robustness of the synthesized results.
        \end{itemize}
        \item[14.] Reporting bias assessment: Describe any methods used to assess risk of bias due to missing results in a synthesis (arising from reporting biases).
        \item[15.] Certainty assessment: Describe any methods used to assess certainty (or confidence) in the body of evidence for an outcome.
    \end{itemize}

    \item \colorbox{lightgray}{\textbf{FINETUNING DATASET}}
    \item[\colorbox{lightgray}{16}] \colorbox{lightgray}{\textbf{Details of the finetuning dataset:}}

        \begin{itemize}
    \item[\colorbox{lightgray}{16a.}] \colorbox{lightgray}{Dataset preprocessing:} Describe the procedures used for processing and preparing academic papers for information extraction.
    \item[\colorbox{lightgray}{16b.}] \colorbox{lightgray}{Dataset format:} Specify the structure of the final dataset, including the format of the input and output data for the LLM.
    \item[\colorbox{lightgray}{16c.}] \colorbox{lightgray}{Data augmentation:} Report any additional datasets used for instruct-finetuning or for increasing the LLM's general domain knowledge.
    \item[\colorbox{lightgray}{16d.}] \colorbox{lightgray}{Dataset curation:} Detail the strategy used to construct the finetuning dataset, including the automation or manual steps used to represent the information from the academic papers.
    \item[\colorbox{lightgray}{16e.}] \colorbox{lightgray}{Dataset composition:} Report the size attributes of the final finetuning dataset.
        \end{itemize}

\item \colorbox{lightgray}{\textbf{LLM FINETUNING}}
    \item[\colorbox{lightgray}{17}] \colorbox{lightgray}{\textbf{Technical finetuning details:}}

    \begin{itemize}
    \item[\colorbox{lightgray}{17a.}] \colorbox{lightgray}{LLM specifications:} Justify the choice of LLM for finetuning, considering model capacity, architectural features, and reported performance on language tasks.
    \item[\colorbox{lightgray}{17b.}] \colorbox{lightgray}{Finetuning strategy:} Discuss the finetuning strategy used, whether classical approach is used requiring a full-model parameter update, or a partial update is employed.
    \item[\colorbox{lightgray}{17c.}] \colorbox{lightgray}{Finetuning settings:} Explain the finetuning procedure, including the adjustment of hyperparameters and the optimization algorithm used for adjusting the model's parameters. Include any techniques used to prevent overfitting, such as dropout, weight decay, or early stopping.
    \item[\colorbox{lightgray}{17d.}] \colorbox{lightgray}{Post finetuning:} Outline any post-finetuning processing steps performed to optimize the LLM's performance.
    \end{itemize}
    
\item \colorbox{lightgray}{\textbf{FINETUNED LLM EVALUATION}}
    \item[\colorbox{lightgray}{18}] \colorbox{lightgray}{\textbf{Validation of the domain-specific LLM:}}

\begin{itemize}
\item[\colorbox{lightgray}{18a.}] \colorbox{lightgray}{LLM benchmarking:} Discuss the initial performance of the LLM before finetuning, on a set of benchmark tasks for which the model will be finetuned.
\item[\colorbox{lightgray}{18b.}] \colorbox{lightgray}{Evaluation stability and reproducibility:} Report any measures taken to ensure the robustness of the evaluation, such as multiple runs with different random seeds and variations in the dataset and the model's initial parameters.
\item[\colorbox{lightgray}{18c.}] \colorbox{lightgray}{Qualitative analysis:} Include an analysis of the types of errors made by the model and potential reasons for these errors.
\item[\colorbox{lightgray}{18d.}] \colorbox{lightgray}{Alignment:} Report the model's performance on the task as well as its ability to produce outputs that are coherent, relevant, and ethically acceptable.
\item[\colorbox{lightgray}{18e.}] \colorbox{lightgray}{Evaluation metrics:} Justify the choice of evaluation metrics based on the task requirements.
\item[\colorbox{lightgray}{18f.}] \colorbox{lightgray}{Qualitative analysis:} Discuss the qualitative analysis of the model's outputs, including case studies of successful and less successful examples.
\end{itemize}

    \item \textbf{RESULTS}
    \begin{itemize}
        \item[19a.] Study selection: Describe the results of the search and selection process, from the number of records identified in the search to the number of studies included in the review, ideally using a flow diagram.
        \item[19b.] Cite studies that might appear to meet the inclusion criteria, but which were excluded, and explain why they were excluded.
        \item[20.] Study characteristics: Cite each included study and present its characteristics.
        \item[21.] Risk of bias in studies: Present assessments of risk of bias for each included study.
        \item[22.] Results of individual studies: For all outcomes, present, for each study: (a) summary statistics for each group (where appropriate) and (b) an effect estimate and its precision (e.g. confidence/credible interval), ideally using structured tables or plots.
        \item[23] \textbf{Results of syntheses:}
        \begin{itemize}
            \item[23a.] Briefly summarise the characteristics and risk of bias among contributing studies.
            \item[23b.] Present results of all statistical syntheses conducted. If meta-analysis was done, present for each the summary estimate and its precision (e.g. confidence/credible interval) and measures of statistical heterogeneity. If comparing groups, describe the direction of the effect.
            \item[23c.] Present results of all investigations of possible causes of heterogeneity among study results.
            \item[23d.] Present results of all sensitivity analyses conducted to assess the robustness of the synthesized results.
        \end{itemize}
        \item[24.] Reporting biases: Present assessments of risk of bias due to missing results (arising from reporting biases) for each synthesis assessed.
        \item[25.] Certainty of evidence: Present assessments of certainty (or confidence) in the body of evidence for each outcome assessed.
    \end{itemize}
    
    \item \textbf{DISCUSSION}
    \begin{itemize}
        \item[26a.] Discussion: Provide a general interpretation of the results in the context of other evidence.
        \item[26b.] Discuss any limitations of the evidence included in the review.
        \item[26c.] Discuss any limitations of the review processes used.
        \item[26d.] Discuss implications of the results for practice, policy, and future research.
\item[\colorbox{lightgray}{26e.}] \colorbox{lightgray}{Discuss processes to enable an ongoing and incremental living systematic review.}
    \end{itemize}
    
    \item \textbf{OTHER INFORMATION}
    \begin{itemize}
        \item[27a.] Registration and protocol: Provide registration information for the review, including register name and registration number, or state that the review was not registered.
        \item[27b.] Indicate where the review protocol can be accessed, or state that a protocol was not prepared.
        \item[27c.] Describe and explain any amendments to the information provided at registration or in the protocol.
        \item[28.] Support: Describe sources of financial or non-financial support for the review, and the role of the funders or sponsors in the review.
        \item[29.] Competing interests: Declare any competing interests of review authors.
        \item[30.] Availability of data, code and other materials: Report which of the following are publicly available and where they can be found: template data collection forms; data extracted from included studies; data used for all analyses; analytic code; any other materials used in the review; \colorbox{lightgray}{availability of the finetuning dataset and the finetuned LLM}.

    \item[\colorbox{lightgray}{31}] \colorbox{lightgray}{\textbf{LLM Legal and Ethical information:}}

            \begin{itemize}
                \item[\colorbox{lightgray}{31.a}] \colorbox{lightgray}{LLM Ethical implications:} Address the potential for the LLM's outputs to cause harm, either through the propagation of biases, the generation of inappropriate content, or the violation of privacy.
                \item[\colorbox{lightgray}{31.b}] \colorbox{lightgray}{LLM Legal implications:} Discuss the legal implications of finetuning LLMs on academic papers, including considerations of copyright laws, fair use, and obtaining permissions from copyright holders where necessary.
                \item[\colorbox{lightgray}{31.c}] \colorbox{lightgray}{LLM Compliance:} Document the process of ensuring compliance with data protection laws and ethical guidelines.
\end{itemize}
        
    \end{itemize}
\end{enumerate}

\section{Discussion}

The development of domain-specific finetuned LLMs for individual disciplines, capable of supporting robust SLRs, presents a promising avenue; however, this pursuit also introduces unique challenges that necessitate further exploration and research.

\subsection{Potential Benefits of PRISMA-DFLLM}

\paragraph{Enhanced Efficiency}

The integration of domain-specific finetuned LLMs into SLRs could significantly enhance the efficiency of the review process. By automating labor-intensive tasks such as data extraction and evidence synthesis, researchers can drastically reduce the time and resources traditionally required for these tasks. For instance, a domain-specific LLM trained on medical literature could automatically extract relevant data from a large number of clinical trials, such as patient demographics, intervention details, and outcome measures, thereby expediting the production of systematic reviews. This increased efficiency could enable researchers to stay abreast of the rapidly expanding body of literature and respond more promptly to emerging research questions.

\paragraph{Scalability and Living Systematic Reviews}

Domain-specific finetuned LLMs offer scalability, facilitating the review of a large volume of literature within a condensed time frame. The capabilities of finetuned language models could be leveraged to analyze a more extensive number of papers at a comprehensive level. This scalability is particularly beneficial in research fields with a high publication rate or when conducting frequent updates of systematic reviews. For example, in the field of infectious diseases, where new research on diseases like COVID-19 is published at a rapid pace, a domain-specific LLM could help researchers keep up with the latest findings. Furthermore, the scalability of domain-specific finetuned LLMs paves the way for the realization of LSRs that incrementally update the knowledge base of the underlying domain-specific LLM. Once the training hyperparameters and data extraction processes have been optimized, the return on the invested effort is then realized through the ability to update the LLM on an ongoing basis.

\paragraph{Discovery of Novel Insights}

The application of PRISMA-DFLLM can potentially uncover novel insights and patterns within the literature. By utilizing specilized LLMs, researchers could identify new connections, trends, and relationships across studies that may not be immediately apparent through traditional review methods. This ability to generate novel insights could enrich the understanding of a research topic, generate new theories and stimulate further investigation.

\paragraph{Dissemination and Collaboration}

A significant benefit of the underlying ideas behind PRISMA-DFLLM is the potential for disseminating finetuned LLMs across different research teams and institutions. Researchers could potentially share the trained models, allowing others to utilize and benefit from their expertise and findings. This dissemination fosters transparency, collaboration, and the building of cumulative knowledge within the research community and sharing the resource overheads in updating the models with new research. For example, teams of researchers could distribute the finetuning tasks among themselves where some focus on dataset curation and others on the LLM finetuning process thus accelerating research.

\subsection{Potential Challenges of Implementing PRISMA-DFLLM}

The implementation of the PRISMA-DFLLM framework, while promising, presents several challenges. These include the automation of data extraction from raw academic articles, the identification of optimal PEFT strategies, ensuring alignment, and the evaluation of finetuned models.

\paragraph{Automating Data Extraction from Raw Academic Articles}

Automating the extraction of data from raw academic articles for the purpose of finetuning LLMs is complex. Academic articles, while generally adhering to certain conventions such as the IMRaD (Introduction, Methods, Results, and Discussion) structure, can vary greatly in their organization and presentation of information. This variability can make it difficult for an LLM to consistently locate and extract the necessary information for systematic reviews, such as study design, participant characteristics, and results. Moreover, academic articles often contain valuable information in non-textual formats such as tables, figures, and images. Extracting and encoding this information as text for LLM training presents another layer of complexity and will likely require additional AI components to solve adequately. Current LLMs are primarily designed to process text and may struggle to interpret and integrate information from these non-textual sources. Developing new methods or tools to automate the extraction and encoding of data from these formats is a pressing research need.

Additionally, the raw data from academic articles are often in PDF format, which is not readily usable for model training. Converting PDFs into a machine-readable format introduces additional steps into the data preparation process, each of which can potentially introduce errors or distortions that require human oversight and verification. Furthermore, the creation of finetuning datasets from academic articles is not a straightforward task. A balance must be struck between providing the model with raw data, which gives it a broader understanding of the content and context of the articles, and providing it with structured data in the form of question-answer pairs, which can guide the model's learning in more specific ways. Determining the optimal balance and integration of these different data types is an open research question.

\paragraph{Optimal PEFT Strategies}

Choosing the best PEFT strategies for developing domain-specific LLMs is a multifaceted task. The decision hinges on the task's nature, the available training data, and computational resources. Task-specific finetuning with PEFT eliminates the need for pretraining but places a premium on effective data curation and optimal hyperparameter training settings. Techniques like Low-Rank Adaptation (LoRA) and Quantized LoRA (QLoRA) offer resource-efficient alternatives, but choosing between them depends on the task's specifics and available resources. An ensemble of specialized models, each finetuned on a task subset, can enhance performance, but coordinating these models and integrating their outputs can be complex and computationally costly. Active learning approaches optimize the finetuning process but require continuous human annotator interaction, introducing potential data privacy, quality control challenges as well as additional overheads. In essence, selecting the most effective finetuning approach requires careful consideration of various factors and potential trade-offs, highlighting the need for continued research in LLM finetuning.

\paragraph{Evaluating Finetuned Models for Alignment}

Securing alignment between the finetuned LLM and the specific task requirements, as well as ethical guidelines, is a complex yet critical aspect of model development. This process extends beyond the technicalities of model training to encompass wider ethical and societal considerations. For instance, the model should be designed to avoid potential biases in its outputs, which requires careful curation and examination of the training data. Additionally, the model should respect privacy norms, especially when dealing with sensitive data or topics. This might involve implementing mechanisms to prevent the model from generating inappropriate or sensitive content. Testing for alignment is a multifaceted process that can employ both quantitative and qualitative methods. Quantitative methods might involve performance metrics such as accuracy, precision, recall, or F1 score, as well as custom benchmarks tailored to the specific task. Qualitative methods, on the other hand, might involve a detailed examination of the model's outputs. For instance, expert reviewers could assess whether the model's responses are contextually appropriate, coherent, and free from bias. They could also evaluate the model's ability to handle complex queries and its sensitivity to the input prompt. Furthermore, alignment testing should also consider the model's interpretability and transparency. For instance, can the model provide explanations for its outputs? Is it clear how the model is using the input data to generate its responses? These are important questions for ensuring the model's alignment with ethical guidelines and user expectations.

\paragraph{Data Availability}

Securing a comprehensive and diverse corpus of academic papers for finetuning LLMs can be a formidable task. The availability of open-access resources can be limited, and the accessibility of certain journals or papers may be hindered by paywalls or licensing agreements. For instance, a researcher aiming to finetune a model on AI literature might discover that key journals in the field are not only concealed behind paywalls, but also explicitly prohibit the use of their content for LLM finetuning. This restriction could significantly limit the diversity and representativeness of the training dataset, thereby impacting the model's ability to specialize sufficiently in a given field and to conduct a full PRISMA review. Furthermore, the issue of data availability is not static but evolves. As new research is published, the training data needs to be updated to ensure the model remains current. This requires ongoing access to new publications, which may not always be guaranteed due to changes in access policies or licensing agreements.

\paragraph{Domain-specific Vocabulary}

Different research domains often employ their own specialized vocabulary and terminology. For example, the term "cell" has different meanings in biology and the context of mobile communication technology. Adapting the language model to accurately understand and generate domain-specific language is crucial, and may require additional finetuning or the use of domain-specific corpora.

\subsection{Future research directions for advancing LLM-enabled literature reviews}

In light of the challenges and opportunities discussed in the previous section, the following table list future research directions under several key categories and attempts to label them according to both the difficulty of the undertaking and its urgency.

{\fontsize{7pt}{10pt}\selectfont 
\begin{longtable}[htbp]{p{2.5cm} p{9cm} c c }
\caption{Future research directions for the PRISMA-DFLLM Framework, highlighting the key categories, difficulty and priority levels of various undertakings. }\\
\hline

\textbf{Category} & \textbf{Research Area} & \textbf{Difficulty} & \textbf{Priority} \\ \hline
\endfirsthead

\multicolumn{4}{c}{{\tablename} \thetable{} -- Continued from previous page} \\
\hline
\textbf{Category} & \textbf{Research Area} & \textbf{Difficulty} & \textbf{Priority} \\ \hline
\endhead
\hline \multicolumn{4}{c}{{Continued on next page}} \\
\endfoot
\hline \hline
\endlastfoot
\multirow{5}{2.5cm}{\centering Automating Data Extraction from Raw Academic Articles for Finetuning Datasets} & Develop and evaluate specialized LLMs for summarization and extracting target textual information from academic articles across various disciplines & Low & Immediate \\ \cline{2-4}
& Investigate the optimal balance and integration of raw data and question-answer pairs in the finetuning dataset & Medium & Immediate \\ \cline{2-4}
& Develop general-purpose tools/libraries for handling data extraction challenges found in the heterogeneity in the structure and content of academic articles across different journals/disciplines & Medium & Immediate \\ \cline{2-4}
& Explore the use of advanced AI agents for extracting textual information from visual and tabular representations in academic articles for automating the curation of finetuning datasets & Difficult & Medium \\ \cline{2-4}
& Assess the feasibility and effectiveness of using unsupervised and semi-supervised learning methods for assisting in data extraction from academic articles & Medium & Long-term \\ \hline

\multirow{9}{2.5cm}{\centering Optimizing Finetuning Strategies and Performance}
& Assess the benefits and limitations of various finetuning strategies, including full-model finetuning and partial finetuning & Medium & Immediate \\ \cline{2-4}
& Develop methods for automating the selection and optimization of finetuning strategies based on the specific task and data characteristics & Medium & Medium \\ \cline{2-4}
& Investigate the use of meta-learning and autoML techniques for optimizing the PEFT process & Difficult & Medium \\ \cline{2-4}
& Investigate the impact of different prompt engineering strategies on the performance of finetuned LLMs & Low & Immediate \\ \cline{2-4}
& Explore the use of iterative and multi-stage finetuning approaches for improving model performance & Medium & Medium \\ \cline{2-4}
& Investigate the impact of different finetuning strategies on LLM performance, including variations in hyperparameters and training approaches & Low & Immediate \\ \cline{2-4}
& Explore techniques for transfer learning and domain adaptation to improve model generalizability across different domains and tasks & Medium & Medium \\ \cline{2-4}
& Assess the benefits of ensemble methods and the combination of multiple finetuned LLMs within the PRISMA-DFLLM framework & Medium & Medium \\ \cline{2-4}
& Investigate strategies to optimize the computational resources required for finetuning and inference processes, ensuring scalability and efficiency & Difficult & Long-term \\ \hline

\multirow{5}{2.5cm}{\centering Evaluating Finetuned Models}
& Compare PRISMA-DFLLM reviews with previous PRISMA reviews in parallel on the same data to benchmark the capabilities of the LLMs  & Low & Immediate \\ \cline{2-4}
& Develop and evaluate task-specific evaluation metrics for assessing the performance of finetuned LLMs in the context of systematic reviews & Low & Immediate \\ \cline{2-4}
& Investigate the use of qualitative and user-centric evaluation methods for assessing model performance & Medium & Medium \\ \cline{2-4}
& Assess the impact of different evaluation strategies on the perceived performance and utility of finetuned LLMs & Medium & Medium \\ \cline{2-4}
& Develop methods for evaluating the robustness and reliability of finetuned LLMs under varying conditions and data characteristics & Medium & Long-term \\ \hline

\multirow{4}{2.5cm}{\centering Ensuring and Testing for Alignment}
& Develop and evaluate methods for testing the alignment of finetuned LLMs with task requirements and ethical guidelines & Low & Immediate \\ \cline{2-4}
& Investigate the use of adversarial testing and bias audits for assessing model alignment & Low & Medium \\ \cline{2-4}
& Develop methods for incorporating feedback from users and stakeholders into the finetuning process to improve model alignment for different disciplines & Difficult & Long-term \\ \cline{2-4}
& Investigate the use of active learning and human-in-the-loop approaches for ensuring model alignment during the finetuning process & Difficult & Long-term \\ \hline

\multirow{4}{2.5cm}{\centering Interpretability and Explainability}
& Enhance the interpretability of finetuned domain-specific LLMs by developing methods to provide insights into the model's decision-making process & Difficult & Immediate \\ \cline{2-4}
& Investigate techniques for uncertainty estimation and sensitivity analysis to quantify the robustness and reliability of the model's predictions & Difficult & Medium \\ \cline{2-4}
& Address biases and misinformation within the framework by exploring methods to identify, mitigate, and provide transparency around potential biases in the training data & Difficult & Immediate \\ \cline{2-4}
& Investigate methods for integrating user feedback loops, allowing researchers to provide input and refine the models based on their domain expertise and preferences & Medium & Medium \\ \hline

\multirow{4}{2.5cm}{\centering Integration, Collaboration and Legal Implications}
& Explore legal ramifications of finetuning LLMs on non-Open Access academic content as well as the options to disseminate both the finetuning datasets and the finetuned LLMs both across different research teams as well as publicly & Medium & Immediate \\ \cline{2-4}
& Explore the integration of external knowledge sources, such as domain-specific ontologies or expert systems, to enhance the model's understanding and reasoning capabilities & Medium & Medium \\ \cline{2-4}
& Investigate methods for collaborative finetuning, knowledge sharing, and model exchange among researchers to foster collaboration and accelerate advancements & Medium & Medium \\ \cline{2-4}
& Examine approaches for integrating the framework with existing research management systems and platforms, ensuring seamless integration with scholarly databases and knowledge repositories & Difficult & Long-term \\ \hline

\end{longtable}
}

\section{Conclusion}

The advent of open-sourced Large Language Models (LLMs) and efficient finetuning techniques heralds a new era in academic research, particularly in the realm of systematic literature reviews (SLRs). The potential of these technologies to revolutionize the way we access knowledge, conduct SLRs, and generate new insights is immense. The proposed methodology and the PRISMA-DFLLM (Domain-specific Finetuned LLMs) framework, which combines the power of expert LLMs with the reporting guidelines of the Preferred Reporting Items for Systematic Reviews and Meta-Analyses (PRISMA), represents a significant stride towards realizing this potential.
These guidelines have been extended from a checklist of 27 to 31 items.

The combination of finetuned LLMs with the PRISMA-DFLLM framework offers the promise of greater efficiency, reusability, and scalability in conducting SLRs. By finetuning LLMs on domain-specific academic papers selected through a principled SLR process, we can create models that are not only more adept at handling specialized fields and applications but also capable of conducting incremental living systematic reviews. This approach democratizes cutting-edge research, empowering researchers across the globe by enabling them to leverage these finetuned models to accelerate advancements in their respective fields.

However, the journey towards fully realizing the potential of PRISMA-DFLLM is not without challenges. From ensuring data availability and automating data extraction from raw academic articles to identifying optimal PEFT strategies and ensuring alignment with task requirements and ethical guidelines, each step presents unique hurdles. Overcoming these challenges requires innovative solutions, ongoing research, and careful consideration of the ethical and societal implications of this work.

This paper has laid out the case for the feasibility of expert, finetuned LLMs to support rigorous SLRs and has outlined the technical requirements for realizing this vision. The proposed extended PRISMA-DFLLM checklist of reporting guidelines provides a roadmap for researchers seeking to implement this approach. As we move forward, it is crucial that we continue to explore, validate, and refine this approach, paving the way for a new era of evidence synthesis and knowledge discovery. The future of academic research is on the horizon, and it is one where AI-enabled SLRs play a new and significant role.

\urlstyle{same}


\begin{thebibliography}{40}
\providecommand{\natexlab}[1]{#1}
\providecommand{\url}[1]{\texttt{#1}}
\expandafter\ifx\csname urlstyle\endcsname\relax
  \providecommand{\doi}[1]{doi: #1}\else
  \providecommand{\doi}{doi: \begingroup \urlstyle{rm}\Url}\fi

\bibitem[ope()]{openai}
{OpenAI}: About.
\newblock \url{https://openai.com/about/}.
\newblock Accessed on 14 June 2023.

\bibitem[Bolton et~al.(2022)Bolton, Hall, Yasunaga, Lee, Manning, and
  Liang]{bolton2022stanford}
E.~Bolton, D.~Hall, M.~Yasunaga, T.~Lee, C.~Manning, and P.~Liang.
\newblock Stanford crfm introduces pubmedgpt 2.7b.
\newblock
  \url{https://hai.stanford.edu/news/stanford-crfm-introduces-pubmedgpt-27b},
  2022.
\newblock Accessed: 13 June 2023.

\bibitem[Brown et~al.(2020)Brown, Mann, Ryder, Subbiah, Kaplan, Dhariwal,
  Neelakantan, Shyam, Sastry, Askell, et~al.]{Brown2020}
T.~B. Brown, B.~Mann, N.~Ryder, M.~Subbiah, J.~Kaplan, P.~Dhariwal,
  A.~Neelakantan, P.~Shyam, G.~Sastry, A.~Askell, et~al.
\newblock Language models are few-shot learners.
\newblock \emph{arXiv preprint arXiv:2005.14165}, 2020.

\bibitem[Chen et~al.(2023)Chen, Zhang, Shi, Li, Smola, and
  Yang]{chen2023parameter}
J.~Chen, A.~Zhang, X.~Shi, M.~Li, A.~Smola, and D.~Yang.
\newblock Parameter-efficient fine-tuning design spaces.
\newblock \emph{arXiv preprint arXiv:2301.01821}, 2023.

\bibitem[Curcic()]{wordsrated2023}
D.~Curcic.
\newblock Number of academic papers published per year.
\newblock
  \url{https://wordsrated.com/number-of-academic-papers-published-per-year}.
\newblock Accessed on 14th June 2023.

\bibitem[Dettmers et~al.(2023)Dettmers, Pagnoni, Holtzman, and
  Zettlemoyer]{dettmers2023qlora}
T.~Dettmers, A.~Pagnoni, A.~Holtzman, and L.~Zettlemoyer.
\newblock Qlora: Efficient finetuning of quantized llms, 2023.

\bibitem[Elliott et~al.(2014)Elliott, Turner, Clavisi, Thomas, Higgins,
  Mavergames, and Gruen]{elliott2017living}
J.~H. Elliott, T.~Turner, O.~Clavisi, J.~Thomas, J.~P. Higgins, C.~Mavergames,
  and R.~L. Gruen.
\newblock Living systematic reviews: an emerging opportunity to narrow the
  evidence-practice gap.
\newblock \emph{PLoS Med}, 11\penalty0 (2):\penalty0 e1001603, 2014.

\bibitem[Forsgren et~al.(2023)Forsgren, Wallstr{\"o}m, Feldthusen, Zechner,
  Sawatzky, and {\"O}hl{\'e}n]{forsgren2023use}
E.~Forsgren, S.~Wallstr{\"o}m, C.~Feldthusen, N.~Zechner, R.~Sawatzky, and
  J.~{\"O}hl{\'e}n.
\newblock The use of text-mining software to facilitate screening of literature
  on centredness in health care.
\newblock \emph{Systematic Reviews}, 12\penalty0 (1):\penalty0 73, 2023.

\bibitem[Gui et~al.(2023)Gui, Ye, and Xiao]{gui2023g}
A.~Gui, J.~Ye, and H.~Xiao.
\newblock G-adapter: Towards structure-aware parameter-efficient transfer
  learning for graph transformer networks.
\newblock \emph{arXiv preprint arXiv:2305.10329}, 2023.

\bibitem[Hu et~al.(2021)Hu, Shen, Wallis, Allen-Zhu, Li, Wang, Wang, and
  Chen]{hu2021lora}
E.~J. Hu, Y.~Shen, P.~Wallis, Z.~Allen-Zhu, Y.~Li, S.~Wang, L.~Wang, and
  W.~Chen.
\newblock Lora: Low-rank adaptation of large language models, 2021.

\bibitem[Hu et~al.(2023)Hu, Lan, Wang, Xu, Lim, Lee, Bing, and
  Poria]{hu2023llm}
Z.~Hu, Y.~Lan, L.~Wang, W.~Xu, E.-P. Lim, R.~K.-W. Lee, L.~Bing, and S.~Poria.
\newblock Llm-adapters: An adapter family for parameter-efficient fine-tuning
  of large language models.
\newblock \emph{arXiv preprint arXiv:2304.01933}, 2023.

\bibitem[Hutton et~al.(2015)Hutton, Salanti, Caldwell, Chaimani, Schmid,
  Cameron, Ioannidis, Straus, Thorlund, et~al.]{hutton2015prisma}
B.~Hutton, G.~Salanti, D.~M. Caldwell, A.~Chaimani, C.~H. Schmid, C.~Cameron,
  J.~P. Ioannidis, S.~Straus, K.~Thorlund, et~al.
\newblock The prisma extension statement for reporting of systematic reviews
  incorporating network meta-analyses of health care interventions: checklist
  and explanations.
\newblock \emph{Annals of internal medicine}, 162\penalty0 (11):\penalty0
  777--784, 2015.

\bibitem[Knafou et~al.(2023)Knafou, Haas, Borissov, Counotte, Low, Imeri,
  Ipekci, Buitrago-Garcia, Heron, Amini, et~al.]{knafou2023ensemble}
J.~Knafou, Q.~Haas, N.~Borissov, M.~Counotte, N.~Low, H.~Imeri, A.~M. Ipekci,
  D.~Buitrago-Garcia, L.~Heron, P.~Amini, et~al.
\newblock Ensemble of deep learning language models to support the creation of
  living systematic reviews for the covid-19 literature.
\newblock \emph{Systematic Reviews}, 12\penalty0 (1):\penalty0 94, 2023.

\bibitem[Kolaski et~al.(2023)Kolaski, Logan, and Ioannidis]{Kolaski2023}
K.~Kolaski, L.~R. Logan, and J.~P.~A. Ioannidis.
\newblock Guidance to best tools and practices for systematic reviews.
\newblock \emph{Systematic Reviews}, 12\penalty0 (1):\penalty0 96, 2023.
\newblock ISSN 2046-4053.
\newblock \doi{10.1186/s13643-023-02255-9}.
\newblock URL \url{https://doi.org/10.1186/s13643-023-02255-9}.

\bibitem[Landhuis(2016)]{landhuis2016scientific}
E.~Landhuis.
\newblock Scientific literature: Information overload.
\newblock \emph{Nature}, 535\penalty0 (7612):\penalty0 457--458, 2016.

\bibitem[Lehman et~al.(2023)Lehman, Hernandez, Mahajan, Wulff, Smith, Ziegler,
  Nadler, Szolovits, Johnson, and Alsentzer]{lehman2023we}
E.~Lehman, E.~Hernandez, D.~Mahajan, J.~Wulff, M.~J. Smith, Z.~Ziegler,
  D.~Nadler, P.~Szolovits, A.~Johnson, and E.~Alsentzer.
\newblock Do we still need clinical language models?
\newblock \emph{arXiv preprint arXiv:2302.08091}, 2023.

\bibitem[Marshall and Wallace(2019)]{marshall2019toward}
I.~J. Marshall and B.~C. Wallace.
\newblock Toward systematic review automation: a practical guide to using
  machine learning tools in research synthesis.
\newblock \emph{Systematic reviews}, 8:\penalty0 1--10, 2019.

\bibitem[Moher et~al.(1999)Moher, Cook, Eastwood, Olkin, Rennie, and
  Stroup]{moher1999improving}
D.~Moher, D.~J. Cook, S.~Eastwood, I.~Olkin, D.~Rennie, and D.~F. Stroup.
\newblock Improving the quality of reports of meta-analyses of randomised
  controlled trials: the quorom statement.
\newblock \emph{The Lancet}, 354\penalty0 (9193):\penalty0 1896--1900, 1999.

\bibitem[Moher et~al.(2009)Moher, Liberati, Tetzlaff, and
  Altman]{moher2009preferred}
D.~Moher, A.~Liberati, J.~Tetzlaff, and D.~G. Altman.
\newblock Preferred reporting items for systematic reviews and meta-analyses:
  the prisma statement.
\newblock \emph{PLoS medicine}, 6\penalty0 (7):\penalty0 e1000097, 2009.

\bibitem[Moher et~al.(2015)Moher, Shamseer, Clarke, Ghersi, Liberati,
  Petticrew, Shekelle, and Stewart]{moher2015preferred}
D.~Moher, L.~Shamseer, M.~Clarke, D.~Ghersi, A.~Liberati, M.~Petticrew,
  P.~Shekelle, and L.~A. Stewart.
\newblock Preferred reporting items for systematic review and meta-analysis
  protocols (prisma-p) 2015 statement.
\newblock \emph{Systematic reviews}, 4\penalty0 (1):\penalty0 1--9, 2015.

\bibitem[Muller et~al.(2023)Muller, Berg, Meneses-Echavez, Ames, Borge, Jardim,
  Cooper, and Rose]{muller2023effect}
A.~E. Muller, R.~C. Berg, J.~F. Meneses-Echavez, H.~M. Ames, T.~C. Borge,
  P.~S.~J. Jardim, C.~Cooper, and C.~J. Rose.
\newblock The effect of machine learning tools for evidence synthesis on
  resource use and time-to-completion: protocol for a retrospective pilot
  study.
\newblock \emph{Systematic Reviews}, 12\penalty0 (1):\penalty0 1--8, 2023.

\bibitem[O’Connor et~al.(2019)O’Connor, Tsafnat, Thomas, Glasziou, Gilbert,
  and Hutton]{o2019question}
A.~M. O’Connor, G.~Tsafnat, J.~Thomas, P.~Glasziou, S.~B. Gilbert, and
  B.~Hutton.
\newblock A question of trust: can we build an evidence base to gain trust in
  systematic review automation technologies?
\newblock \emph{Systematic reviews}, 8\penalty0 (1):\penalty0 1--8, 2019.

\bibitem[Page et~al.(2021)Page, McKenzie, Bossuyt, Boutron, Hoffmann, Mulrow,
  Shamseer, Tetzlaff, Akl, Brennan, et~al.]{page2021prisma}
M.~J. Page, J.~E. McKenzie, P.~M. Bossuyt, I.~Boutron, T.~C. Hoffmann, C.~D.
  Mulrow, L.~Shamseer, J.~M. Tetzlaff, E.~A. Akl, S.~E. Brennan, et~al.
\newblock The prisma 2020 statement: an updated guideline for reporting
  systematic reviews.
\newblock \emph{International journal of surgery}, 88:\penalty0 105906, 2021.

\bibitem[Qureshi et~al.(2023)Qureshi, Shaughnessy, Gill, Robinson, Li, and
  Agai]{qureshi2023chatgpt}
R.~Qureshi, D.~Shaughnessy, K.~A. Gill, K.~A. Robinson, T.~Li, and E.~Agai.
\newblock Are chatgpt and large language models “the answer” to bringing us
  closer to systematic review automation?
\newblock \emph{Systematic Reviews}, 12\penalty0 (1):\penalty0 72, 2023.

\bibitem[Sarkis-Onofre et~al.(2021)Sarkis-Onofre, Catal{\'a}-L{\'o}pez,
  Aromataris, and Lockwood]{sarkis2021properly}
R.~Sarkis-Onofre, F.~Catal{\'a}-L{\'o}pez, E.~Aromataris, and C.~Lockwood.
\newblock How to properly use the prisma statement.
\newblock \emph{Systematic Reviews}, 10\penalty0 (1):\penalty0 1--3, 2021.

\bibitem[Shea et~al.(2017)Shea, Reeves, Wells, Thuku, Hamel, Moran, Moher,
  Tugwell, Welch, Kristjansson, et~al.]{shea2017amstar}
B.~J. Shea, B.~C. Reeves, G.~Wells, M.~Thuku, C.~Hamel, J.~Moran, D.~Moher,
  P.~Tugwell, V.~Welch, E.~Kristjansson, et~al.
\newblock Amstar 2: a critical appraisal tool for systematic reviews that
  include randomised or non-randomised studies of healthcare interventions, or
  both.
\newblock \emph{bmj}, 358, 2017.

\bibitem[Sheng et~al.(2021)Sheng, Chang, Natarajan, and
  Peng]{sheng2021societal}
E.~Sheng, K.-W. Chang, P.~Natarajan, and N.~Peng.
\newblock Societal biases in language generation: Progress and challenges.
\newblock \emph{arXiv preprint arXiv:2105.04054}, 2021.

\bibitem[Singhal et~al.(2022)Singhal, Azizi, Tu, Mahdavi, Wei, Chung, Scales,
  Tanwani, Cole-Lewis, Pfohl, et~al.]{singhal2022large}
K.~Singhal, S.~Azizi, T.~Tu, S.~S. Mahdavi, J.~Wei, H.~W. Chung, N.~Scales,
  A.~Tanwani, H.~Cole-Lewis, S.~Pfohl, et~al.
\newblock Large language models encode clinical knowledge.
\newblock \emph{arXiv preprint arXiv:2212.13138}, 2022.

\bibitem[Singhal et~al.(2023)Singhal, Tu, Gottweis, Sayres, Wulczyn, Hou,
  Clark, Pfohl, Cole-Lewis, Neal, et~al.]{singhal2023towards}
K.~Singhal, T.~Tu, J.~Gottweis, R.~Sayres, E.~Wulczyn, L.~Hou, K.~Clark,
  S.~Pfohl, H.~Cole-Lewis, D.~Neal, et~al.
\newblock Towards expert-level medical question answering with large language
  models.
\newblock \emph{arXiv preprint arXiv:2305.09617}, 2023.

\bibitem[Stevens et~al.(2018)Stevens, Garritty, Hersi, and
  Moher]{stevens2018developing}
A.~Stevens, C.~Garritty, M.~Hersi, and D.~Moher.
\newblock Developing prisma-rr, a reporting guideline for rapid reviews of
  primary studies (protocol).
\newblock \emph{Equator Network}, 2018.

\bibitem[Stewart et~al.(2015)Stewart, Clarke, Rovers, Riley, Simmonds, Stewart,
  and Tierney]{stewart2015preferred}
L.~A. Stewart, M.~Clarke, M.~Rovers, R.~D. Riley, M.~Simmonds, G.~Stewart, and
  J.~F. Tierney.
\newblock Preferred reporting items for systematic review and meta-analyses of
  individual participant data: the prisma-ipd statement.
\newblock \emph{Jama}, 313\penalty0 (16):\penalty0 1657--1665, 2015.

\bibitem[Stroup et~al.(2000)Stroup, Berlin, Morton, Olkin, Williamson, Rennie,
  Moher, Becker, Sipe, Thacker, et~al.]{stroup2000meta}
D.~F. Stroup, J.~A. Berlin, S.~C. Morton, I.~Olkin, G.~D. Williamson,
  D.~Rennie, D.~Moher, B.~J. Becker, T.~A. Sipe, S.~B. Thacker, et~al.
\newblock Meta-analysis of observational studies in epidemiology: a proposal
  for reporting.
\newblock \emph{Jama}, 283\penalty0 (15):\penalty0 2008--2012, 2000.

\bibitem[Touvron et~al.(2023)Touvron, Lavril, Izacard, Martinet, Lachaux,
  Lacroix, Rozi{\`e}re, Goyal, Hambro, Azhar, et~al.]{touvron2023llama}
H.~Touvron, T.~Lavril, G.~Izacard, X.~Martinet, M.-A. Lachaux, T.~Lacroix,
  B.~Rozi{\`e}re, N.~Goyal, E.~Hambro, F.~Azhar, et~al.
\newblock Llama: Open and efficient foundation language models.
\newblock \emph{arXiv preprint arXiv:2302.13971}, 2023.

\bibitem[Tricco et~al.(2018)Tricco, Lillie, Zarin, O'Brien, Colquhoun, Levac,
  Moher, Peters, Horsley, Weeks, et~al.]{tricco2018prisma}
A.~C. Tricco, E.~Lillie, W.~Zarin, K.~K. O'Brien, H.~Colquhoun, D.~Levac,
  D.~Moher, M.~D. Peters, T.~Horsley, L.~Weeks, et~al.
\newblock Prisma extension for scoping reviews (prisma-scr): checklist and
  explanation.
\newblock \emph{Annals of internal medicine}, 169\penalty0 (7):\penalty0
  467--473, 2018.

\bibitem[Tsafnat et~al.(2014)Tsafnat, Glasziou, Choong, Dunn, Galgani, and
  Coiera]{tsafnat2014systematic}
G.~Tsafnat, P.~Glasziou, M.~K. Choong, A.~Dunn, F.~Galgani, and E.~Coiera.
\newblock Systematic review automation technologies.
\newblock \emph{Systematic reviews}, 3:\penalty0 1--15, 2014.

\bibitem[Wang et~al.(2022)Wang, Agarwal, Mukherjee, Liu, Gao, Awadallah, and
  Gao]{wang2022adamix}
Y.~Wang, S.~Agarwal, S.~Mukherjee, X.~Liu, J.~Gao, A.~H. Awadallah, and J.~Gao.
\newblock Adamix: Mixture-of-adaptations for parameter-efficient model tuning.
\newblock \emph{arXiv preprint arXiv:2210.17451}, 2022.

\bibitem[Whiting et~al.(2016)Whiting, Savovi{\'c}, Higgins, Caldwell, Reeves,
  Shea, Davies, Kleijnen, Churchill, et~al.]{whiting2016robis}
P.~Whiting, J.~Savovi{\'c}, J.~P. Higgins, D.~M. Caldwell, B.~C. Reeves,
  B.~Shea, P.~Davies, J.~Kleijnen, R.~Churchill, et~al.
\newblock Robis: a new tool to assess risk of bias in systematic reviews was
  developed.
\newblock \emph{Journal of clinical epidemiology}, 69:\penalty0 225--234, 2016.

\bibitem[Wu et~al.(2023)Wu, Irsoy, Lu, Dabravolski, Dredze, Gehrmann, Kambadur,
  Rosenberg, and Mann]{wu2023bloomberggpt}
S.~Wu, O.~Irsoy, S.~Lu, V.~Dabravolski, M.~Dredze, S.~Gehrmann, P.~Kambadur,
  D.~Rosenberg, and G.~Mann.
\newblock Bloomberggpt: A large language model for finance.
\newblock \emph{arXiv preprint arXiv:2303.17564}, 2023.

\bibitem[Zhou et~al.(2023)Zhou, Liu, Xu, Iyer, Sun, Mao, Ma, Efrat, Yu, Yu,
  et~al.]{zhou2023lima}
C.~Zhou, P.~Liu, P.~Xu, S.~Iyer, J.~Sun, Y.~Mao, X.~Ma, A.~Efrat, P.~Yu, L.~Yu,
  et~al.
\newblock Lima: Less is more for alignment.
\newblock \emph{arXiv preprint arXiv:2305.11206}, 2023.

\bibitem[Zorzela et~al.(2016)Zorzela, Loke, Ioannidis, Golder, Santaguida,
  Altman, Moher, Vohra, et~al.]{zorzela2016prisma}
L.~Zorzela, Y.~K. Loke, J.~P. Ioannidis, S.~Golder, P.~Santaguida, D.~G.
  Altman, D.~Moher, S.~Vohra, et~al.
\newblock Prisma harms checklist: improving harms reporting in systematic
  reviews.
\newblock \emph{bmj}, 352, 2016.

\end{thebibliography}

\appendix

\end{document}